\documentclass{article}

\usepackage[final]{main}
\usepackage{graphicx}
\usepackage{times}
\setcitestyle{numbers,square,comma}
\usepackage{multirow,tabularx}
\usepackage{enumitem}
\newcolumntype{Y}{>{\centering\arraybackslash}X}
\usepackage[utf8]{inputenc} % allow utf-8 input
\usepackage[T1]{fontenc}    % use 8-bit T1 fonts
\usepackage{hyperref}       % hyperlinks
\usepackage{url}            % simple URL typesetting
\usepackage{booktabs}       % professional-quality tables
\usepackage{amsfonts}       % blackboard math symbols
\usepackage{nicefrac}       % compact symbols for 1/2, etc.
\usepackage{microtype}      % microtypography
\usepackage{xcolor}         % colors

\title{Transfer Learning Approaches for Knowledge Discovery in Grid-based Geo-Spatiotemporal Data}

\author{%
  Aishwarya Sarkar\\
  Department of Computer Science\\
  Iowa State University\\
  Ames, IA 50011 \\
  \texttt{asarkar1@iastate.edu} \\
   \And
   Jien Zhang \\
   Department of Ecology, Evolution, and Organismal Biology\\
   Iowa State University \\
    Ames, IA 50011 \\
   \texttt{jienz@iastate.edu} \\
   \And
   Chaoqun Lu \\
   Department of Ecology, Evolution, and Organismal Biology\\
   Iowa State University \\
    Ames, IA 50011 \\
   \texttt{clu@iastate.edu} \\
   \And
   Ali Jannesari \\
    Department of Computer Science\\
  Iowa State University\\
  Ames, IA 50011 \\
   \texttt{jannesar@iastate.edu} \\
}

\begin{document}

\maketitle

\begin{abstract}
Extracting and meticulously analyzing geo-spatiotemporal features is crucial to recognize intricate underlying causes of natural events, such as floods. Limited evidence about hidden factors leading to climate change makes it challenging to predict regional water discharge accurately. In addition, the explosive growth in complex geo-spatiotemporal environment data that requires repeated learning by the state-of-the-art neural networks for every new region emphasizes the need for new computationally efficient methods, advanced computational resources, and extensive training on a massive amount of available monitored data. We, therefore, propose HydroDeep, an effectively reusable pretrained model to address this problem of transferring knowledge from one region to another by effectively capturing their intrinsic geo-spatiotemporal variance. Further, we present four transfer learning approaches on HydroDeep for spatiotemporal interpretability that improve Nash–Sutcliffe efficiency by $9\%$ to $108\%$ in new regions with a $95\%$ reduction in time.
\end{abstract}
\section{Introduction}
Spatiotemporal variance influencing local water discharge is inevitable due to varying soil property, climate, and land usage among different regions. Training individual hydrological models to capture regional spatiotemporal features requires extensive training and computational resources. These regions, although varying in local characteristics, still follow fundamental hydrological dependencies. Transfer learning is a well-known solution to reuse a trained model to reduce training duration for a new dataset.  Although most commonly applied in computer vision \cite{tan2018survey,celik2020automated,pires2020convolutional} and time-series prediction \cite{ma2019improving,gupta2020transfer,lampos2021tracking}, recent works \cite{goswami2020transfer,chakraborty2021transfer} in other application domains show that these techniques have also been effective in knowledge guided neural networks that are powered by process-based (PB) mechanisms. In hydrology, PB  models that rely on domain-specific principles and mathematical formulations, although often criticized as overly complex are elegant in describing large-scale patterns \cite{kleidon2009thermodynamics,wang2009model} where knowledge of distributed state variables and physical constraints is essential. However, understanding a system's general organization does not provide insights into how the principal variables interact over space and time. \cite{fatichi2016overview} discuss all the prevalent challenges of distributed hydrological PB models. On the other hand, several works show an increase in deep learning (DL) applications in hydrology. A recent survey of 129 publications finds the use of various state-of-the-art DL network architectures, namely, convolutional neural networks (CNNs), long short-term memory networks (LSTMs), and gated recurrent units (GRUs) \cite{sit2020comprehensive}. However, limited knowledge of underlying PB mechanisms in DL algorithms makes PB models irreplaceable in studying environmental systems. As a result, knowledge-guided deep neural networks are an active research area in hydrology \cite{yang2019real,WANG2020124700,jia2021physics} as well as other domains \cite{karpatne2017physics,ajay2018augmenting,masi2021thermodynamics}. 

In this paper, we present HydroDeep that couples a PB hydro-ecological model with a combination of one-dimensional CNN and LSTM, to capture regional grid-based geo-spatiotemporal features of a watershed contributing to water discharge that influence an event of flood. The combination of CNN and LSTM that our PB-DL based network uses showed promising results in computer vision \cite{donahue2015long}, speech recognition and natural language processing \cite{vinyals2015show}, and other time-series analysis \cite{Huang_2018}. In our experiments, HydroDeep outperformed the Nash–Sutcliffe efficiency \cite{moriasi2007model} of standalone CNNs and LSTMs by 1.6\% and 10.5\% respectively. We further propose a new application area of transfer learning to analyze similarities and dissimilarities in geo-spatiotemporal characteristics of watersheds while reducing the extensive training time required in training local DL models. Recent research in hydrology shows transfer learning has been applied in predicting water temperature in unmonitored lakes \cite{willard2020predicting} and water quality prediction system \cite{chen2021transfer}. In \cite{kimura2020convolutional} transfer learning was applied for flood prediction, similar to us. However, they converted their time-series data to images to use transfer learning in a purely CNN architecture for time-series prediction with a reduction in computational cost by $18\%$. In this paper, we explore how transferrable geo-spatiotemporal features are when kept in their original spatiotemporal form.
\section{Approach}
\subsection{HydroDeep}
If ~$g_i\in\{g_1, g_2, \dots, g_L\}$ is the spatial grid vector where L is the total number of grids covering a region having spatial coordinates ~$c_{g_i}\in\{(x_{g_1},y_{g_1}), (x_{g_2},y_{g_2}), \dots, (x_{g_L},y_{g_L})\}$, the distance to these grids from the nearest river or water source is ~$d_{g_i}\in\{d_{g_1}, d_{g_2}, \dots, d_{g_L}\}$. On a certain day $t$, the grids have precipitation measurements $p_{g_i}$ where $p_{g_i}\in\{p_{g_1}, p_{g_2}, \dots , p_{g_L}\}$ which are mapped to their corresponding grid-based PB simulated runoff, $r_{g_i}$ where $r_{g_i}\in\{r_{g_1}, r_{g_1}, \dots , r_{g_L}\}$. The extent to which each grid's precipitation contributes to the river discharge  depends  extensively on  the  grid’s  distance  to  the nearest water source. The distance vector is thus transformed to a distance weight vector $\widetilde{d}_{g_i}$ such that higher distance weights are applied to $p_{g_i}$ if the said grid is closer to a local river thus contributing more to the regional river discharge (Appendix Section 1). We denote weighted precipitation vector as $\widetilde{p}_{g_i}=\widetilde{d}_i \odot {p}_{g_i}$. On day t, the input vectors $\widetilde{p}_{g_i, t}$ and $r_{g_i, t}$ are mapped with respective daily river discharge observations $D_t \in \{D_1, D_2, \dots, D_T\}$ where $T$ is the total number of daily river discharge observations. Input vector $x_t$ can be shown as $x_t = f[(\widetilde{p}_{g_i, t}), (r_{g_i, t})]$ where $i\in \{1, 2, \dots,L\}$ and $t\in \{1, 2, \dots, T\}$. From a multivariate time-series point of view, we denote our inputs as $X=(x_1, x_2, \dots, x_T)\in\mathbb{R}^{n\times{T}}$ where $x_t\in\mathbb{R}^n$ and $n$ denotes the total number of input variables per day which in our case is $2L$. We want to predict the corresponding target outputs $D_t$. The aim is to obtain a non-linear mapping between $X$ and $D$. The motive behind the integration of CNN and LSTM in HydroDeep lies in capturing both the spatial and temporal dependencies of a watershed. The CNN layers help extract local geospatial features between the input variables and pass them to LSTMs to support temporal sequence prediction. HydroDeep has an initial input layer customizable to different input shapes (number of grids) to make our model easily transferable to other watersheds. If $t_p$ denotes the prediction day, HydroDeep trains on inputs from $(t_p - 7)$ to $(t_p-1)$, a weekly time window validated by empirical studies, to predict the output of day $t_p$. A continuous drought for days can be followed by a hurricane, leading to a flood overnight, making it crucial to include the target day $t_p$'s inputs as the second input to our network, which is concatenated with the last LSTM layer's output, and are processed collectively by a fully connected layer. We use data from Jan 1, 2000, to Dec 31, 2011, for training and Jan 1, 2012, to Dec 31, 2016, for evaluation. We optimized the hyperparameters of HydroDeep by random search. (Appendix Section 2)

\subsection{Transfer Learning}An event of flood or drought is heavily dependent on environmental drivers. Likewise, each of these drivers' inherent local spatiotemporal patterns is bound to be unique based on their geographical location. Consequently, one model that has learned the local spatiotemporal patterns of a region will fail to perform accurately for a geographically distant region with different characteristics. As a result, the model should be retrained perpetually for every new unique region. Alternatively, a global model can be trained on a larger area covering many watersheds to address this problem, but it will fail to capture the local patterns. Besides, both these methods are expensive as they require more training time and computational power to train such a model of global extent. Therefore, we use transfer learning to reuse HydroDeep's knowledge from one region to another. More formally, transfer learning consists of a domain $\widetilde{D}$ and a task $\widetilde{T}$ where the domain $\widetilde{D}$ is the marginal probability distribution $P(X)$ over an input feature space $X = \{x_1, x_2,\dots, x_N\}$ where $N$ is the total number of input features. Given a domain $\widetilde{D} = \{X, P(X)\}$, a task $\widetilde{T}$ consists of a conditional probability distribution $P(Y|X)$ over a label space $Y$. The conditional probability distribution is usually learned from the pairs $\{x_i, y_i\}$ in the training samples where $x_i \in X$ and $y_i \in Y$. Suppose there is a source domain $\widetilde{D}_{source}$ with a source task $\widetilde{T}_{source}$ and a target domain $\widetilde{D}_{target}$ with a target task $\widetilde{T}_{target}$, through transfer learning we try to learn the target conditional probability distribution $P(Y_{target}| X_{target})$ in $\widetilde{D}_{target}$, from the knowledge learned from $\widetilde{D}_{source}$ and $\widetilde{T}_{target}$ \cite{laptev2018applied}.

\section{Experimental Design}
\subsection{Dataset and the Process-based Model - DLEM}
For experiments, we use a  PB hydro-ecological model named the Dynamic Land Ecosystem Model (DLEM) that mimics the plant physiological, biogeochemical, and hydrological processes in the plant-soil-water-river continuum \cite{liu-et-al:scheme,lu2013net:kl-one}. In DLEM, the design of grid-to-grid connection tracks significant features of a region, including within-grid heterogeneity, grid-to-grid flow, and land-aquatic linkage. DLEM models the water movement from land to aquatic systems at a daily time step with each grid cell comprising multiple land cover types, rivers, and lakes with their area percentage prescribed by land-use history data \cite{Lu2020}. Research shows that DLEM has been extensively validated against measurements from LTER, NEON, Ameri Flux, USDA crop yield survey, and USGS gauge monitoring and are widely used to quantify the spatiotemporal variations in the pool and fluxes of water, carbon, and nitrogen coupling (water-C-N) at the site, and regional scales \cite{liu-et-al:scheme,lu2013net:kl-one,yang2015increased,yu2019largely,lu2019-et-al:scheme}. The preliminary results from DLEM at the outlet of the Mississippi and Atchafalaya river basin (MARB) show that the variations of daily river discharge are very close to the USGS observed river discharge over the years \cite{Lu2020}.

We pulled dataset from the U.S. Geological Survey's (USGS) daily discharge measurements for Iowa Streams having daily discharge measurements of 23 watersheds covering the state of Iowa \cite{jones-et-al:scheme}. Alongside, we use daily precipitation at a 5-arc-min resolution generated from high-resolution gridded meteorological data products from station observations by the Climatic Research Unit (CRU) of the University of East Anglia \cite{doi:10.1002/joc.1181}, and North America Regional Reanalysis (NARR) dataset from a combination of modeled and observed data \cite{10.1175/BAMS-87-3-343}. We also use DLEM-simulated surface and subsurface runoff to guide our DL network \cite{Lu2020}. Climate, land management, and environmental drivers steer DLEM simulation and are used to represent our ``best estimate'' of land-to-aquatic surface and subsurface runoff across the watershed. For our experiment, we chose the Thompson Fork Grand River basin at David City (watershed 13) as our source domain due to its smaller size and transferred its knowledge to 5 target domains - West Nodaway River  near Shambaugh (watershed 14), East Nishnabotna River near Shenandoah (watershed 15), Turkey River near Garber (watershed 4), South Skunk River near Oskaloosa (watershed 10), and Rock River near Hawarden (watershed 23) \cite{jones-et-al:scheme}.  Watershed 13 (w13) has 29 grids with each grid having two input time-series signals corresponding to precipitation and DLEM runoff. Daily river discharge observations are given as labels to all 29 grids.
\subsection{Regional Knowledge Transfer for Spatiotemporal Analysis}
We have experimented with four transfer learning approaches to reuse HydroDeep’s knowledge in predicting river discharge in distant watersheds (Figure 1). The goal was to find the best approach to transfer HydroDeep’s grid-based spatiotemporal knowledge from one watershed to another to reduce the required training iterations on a new region and interpret the geo-spatiotemporal similarities and dissimilarities between the source and the target. In the first approach, we preserve the original HydroDeep’s spatiotemporal knowledge from the source w13 and use it solely in T-HydroDeep-1 (T-HD-1) to test the new target watersheds. Secondly, we transfer the original HydroDeep’s spatiotemporal knowledge in T-HydroDeep-2 (T-HD-2) and allow it to finetune on the target. In the third and fourth approaches, we take turns in finetuning just the temporal features in T-HydroDeep-3 (T-HD-3) and the spatial features in T-HydroDeep-4 (T-HD-4). The CNN layers and the LSTM layers are responsible for learning spatial and temporal features, respectively. When we finetune one kind of layer, we freeze the other to keep the originally learned features intact. Table 1 shows the observations from our ablation study.
\begin{figure}
\centering
\includegraphics[width=0.9\textwidth]{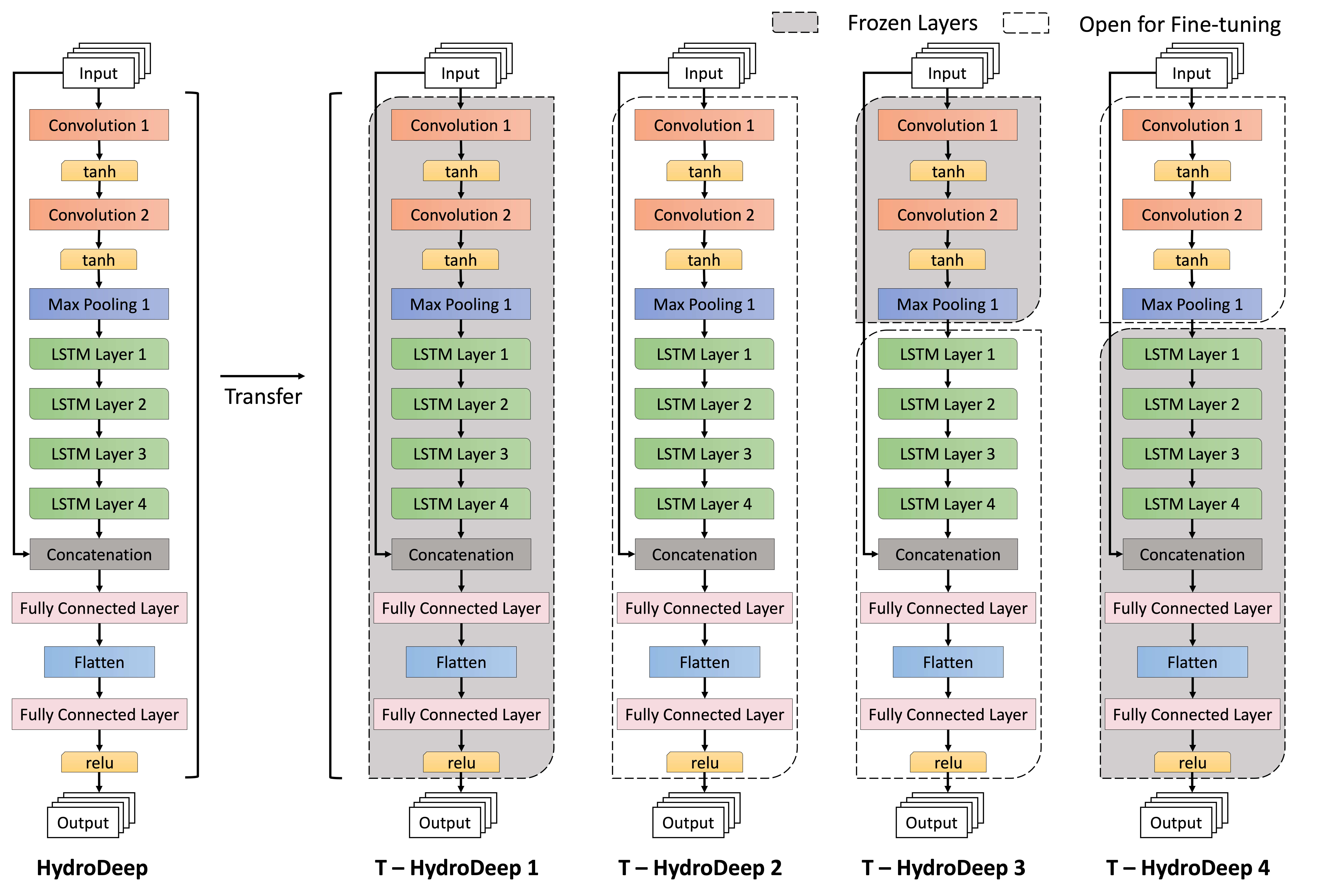} % Reduce the figure size so that it is slightly narrower than the column.
\caption{Four transfer learning approaches in reusing HydroDeep's layers for a new target.}
\label{fig2}
\end{figure}
\section{Results}The watersheds w14 and w15, although being adjacent to the source w13 \cite{jones-et-al:scheme}, are observed to have distinct spatial features as T-HD-4 proved to be the best approach for both the watersheds. We also observed that the second-best approach for w14 is T-HD-1, which means w13 and w14 have similar temporal features, but since they vary in spatial features, only the spatial layers needed to be finetuned (T-HD-4) to improve prediction accuracy. Similarly, the result shows that w15 has more distinct spatial features than temporal features from w13, as T-HD-2 shows the second-best performance, just behind T-HD-4. However, w4 and w10, both being far from w13, show distinct spatiotemporal features as T-HD-2 shows the best performance. Note that w13 has only 29 spatial grids, whereas both w4 and w10 have 61 and 65 grids, respectively. HydroDeep's knowledge is transferred to targets almost double the area of the source and still achieved the best performances among all the watersheds included in our experiment. This supports our arguments that more data availability will increase HydroDeep's performance; transfer learning approaches work in targets larger in area than the source and can be used to analyze watersheds' spatiotemporal characteristics. The original HydroDeep was pretrained for 300 iterations on w13 in $\sim3138$ seconds to achieve 0.63 NSE outperforming the baseline neural network architectures - CNN by $1.6\%$, LSTM by $10.5\%$, GRU by $26\%$, and Bi-directional LSTM by $53.6\%$ that are commonly used for hydrological modeling (Appendix Section 3). In contrast, the transferred models on the new targets (w14, w15, w4, w10, and w23) achieved significant performance in just 20 training iterations in $\sim155$ seconds, a $\sim95\%$ reduction in time. In Table 1, training HydroDeep (HD) from scratch in each target watershed remains low when run for 20 iterations. Thereby, using transfer learning, the prediction performance on the individual watersheds increased by $44\%$ in w14, $108\%$ in w15, $15\%$ in w4, $9\%$ in w10, and $39\%$ in w23 in terms of NSE. 

\begin{table*}[h]
\small
\centering
\caption{Performance comparison of untrained HydroDeep and four transfer learning approaches where HydroDeep was pretrained on w13 with 29 grids before being transferred to other watersheds of varying distance and size. (Appendix Section 4)}
\setlength\tabcolsep{5pt} % default value: 6pt
\begin{tabular}{lrrrrrrrr}
    \toprule
        \textbf{Target Watersheds} &	\textbf{No. of grids}& \textbf{HD} &	\textbf{T-HD-1} & \textbf{T-HD-2} & \textbf{T-HD-3} & \textbf{T-HD-4} & \textbf{Time}\\
        {} & {} & (NSE) & (NSE) & (NSE) & (NSE) & (NSE) & (seconds)\\
    \midrule
        Watershed 14 (w14) & 34 & 0.27 &	0.33 &	0.30 &	0.32  &	\textbf{0.39} & 154.35 $\pm$ 20.81\\
        Watershed 15 (w15) & 39 & 0.24 & 0.46 & 0.47 & 0.42 &	\textbf{0.50} & 154.87 $\pm$ 20.38\\
        Watershed 4 (w4) &	61 & 0.71 &	0.76 &	\textbf{0.82} &	0.81 &	0.82 & 156.52 $\pm$ 20.44\\
        Watershed 10 (w10) & 65 & 0.80 &	0.82 &	\textbf{0.87} &	0.86 &	0.86 & 157.15 $\pm$ 21.40\\
        Watershed 23 (w23) & 32 & 0.36 &	0.45 &	0.38 &	\textbf{0.50} &	0.46 & 154.21 $\pm$ 20.84\\
    \bottomrule
\end{tabular}
\label{table1}
\end{table*}
\section{Conclusion and Future Work}
This paper illustrates a new application of transfer learning techniques in interpreting geo-spatiotemporal characteristics of watersheds with limited computational resources and a $95\%$ reduction in time. We believe that a smaller grid-scale resolution will help HydroDeep to capture local features on a finer scale. In the future, we will run our experiments on more watersheds to better quantify the performance of HydroDeep and its variants. Further, we plan to perform extensive research on how to select the source watershed(s) more effectively. 
\bibliographystyle{unsrt}
\bibliography{main}

\begin{thebibliography}{10}

\bibitem{tan2018survey}
Chuanqi Tan, Fuchun Sun, Tao Kong, Wenchang Zhang, Chao Yang, and Chunfang Liu.
\newblock A survey on deep transfer learning.
\newblock In {\em International conference on artificial neural networks},
  pages 270--279. Springer, 2018.

\bibitem{celik2020automated}
Yusuf Celik, Muhammed Talo, Ozal Yildirim, Murat Karabatak, and U~Rajendra
  Acharya.
\newblock Automated invasive ductal carcinoma detection based using deep
  transfer learning with whole-slide images.
\newblock {\em Pattern Recognition Letters}, 133:232--239, 2020.

\bibitem{pires2020convolutional}
Rafael Pires~de Lima and Kurt Marfurt.
\newblock Convolutional neural network for remote-sensing scene classification:
  Transfer learning analysis.
\newblock {\em Remote Sensing}, 12(1):86, 2020.

\bibitem{ma2019improving}
Jun Ma, Jack~CP Cheng, Changqing Lin, Yi~Tan, and Jingcheng Zhang.
\newblock Improving air quality prediction accuracy at larger temporal
  resolutions using deep learning and transfer learning techniques.
\newblock {\em Atmospheric Environment}, 214:116885, 2019.

\bibitem{gupta2020transfer}
Priyanka Gupta, Pankaj Malhotra, Jyoti Narwariya, Lovekesh Vig, and Gautam
  Shroff.
\newblock Transfer learning for clinical time series analysis using deep neural
  networks.
\newblock {\em Journal of Healthcare Informatics Research}, 4(2):112--137,
  2020.

\bibitem{lampos2021tracking}
Vasileios Lampos, Maimuna~S Majumder, Elad Yom-Tov, Michael Edelstein, Simon
  Moura, Yohhei Hamada, Molebogeng~X Rangaka, Rachel~A McKendry, and Ingemar~J
  Cox.
\newblock Tracking covid-19 using online search.
\newblock {\em NPJ digital medicine}, 4(1):1--11, 2021.

\bibitem{goswami2020transfer}
Somdatta Goswami, Cosmin Anitescu, Souvik Chakraborty, and Timon Rabczuk.
\newblock Transfer learning enhanced physics informed neural network for
  phase-field modeling of fracture.
\newblock {\em Theoretical and Applied Fracture Mechanics}, 106:102447, 2020.

\bibitem{chakraborty2021transfer}
Souvik Chakraborty.
\newblock Transfer learning based multi-fidelity physics informed deep neural
  network.
\newblock {\em Journal of Computational Physics}, 426:109942, 2021.

\bibitem{kleidon2009thermodynamics}
A~Kleidon, Stan Schymanski, and M~Stieglitz.
\newblock Thermodynamics, irreversibility, and optimality in land surface
  hydrology.
\newblock In {\em Bioclimatology and natural hazards}, pages 107--118.
  Springer, 2009.

\bibitem{wang2009model}
J~Wang and Rafael~L Bras.
\newblock A model of surface heat fluxes based on the theory of maximum entropy
  production.
\newblock {\em Water resources research}, 45(11), 2009.

\bibitem{fatichi2016overview}
Simone Fatichi, Enrique~R Vivoni, Fred~L Ogden, Valeriy~Y Ivanov, Benjamin
  Mirus, David Gochis, Charles~W Downer, Matteo Camporese, Jason~H Davison,
  Brian Ebel, et~al.
\newblock An overview of current applications, challenges, and future trends in
  distributed process-based models in hydrology.
\newblock {\em Journal of Hydrology}, 537:45--60, 2016.

\bibitem{sit2020comprehensive}
Muhammed Sit, Bekir~Z Demiray, Zhongrun Xiang, Gregory~J Ewing, Yusuf Sermet,
  and Ibrahim Demir.
\newblock A comprehensive review of deep learning applications in hydrology and
  water resources.
\newblock {\em Water Science and Technology}, 2020.

\bibitem{yang2019real}
Shuyu Yang, Dawen Yang, Jinsong Chen, and Baoxu Zhao.
\newblock Real-time reservoir operation using recurrent neural networks and
  inflow forecast from a distributed hydrological model.
\newblock {\em Journal of Hydrology}, 579:124229, 2019.

\bibitem{WANG2020124700}
Nanzhe Wang, Dongxiao Zhang, Haibin Chang, and Heng Li.
\newblock Deep learning of subsurface flow via theory-guided neural network.
\newblock {\em Journal of Hydrology}, 584:124700, 2020.

\bibitem{jia2021physics}
Xiaowei Jia, Jared Willard, Anuj Karpatne, Jordan~S Read, Jacob~A Zwart,
  Michael Steinbach, and Vipin Kumar.
\newblock Physics-guided machine learning for scientific discovery: An
  application in simulating lake temperature profiles.
\newblock {\em ACM/IMS Transactions on Data Science}, 2(3):1--26, 2021.

\bibitem{karpatne2017physics}
Anuj Karpatne, William Watkins, Jordan Read, and Vipin Kumar.
\newblock Physics-guided neural networks (pgnn): An application in lake
  temperature modeling.
\newblock {\em arXiv preprint arXiv:1710.11431}, 2017.

\bibitem{ajay2018augmenting}
Anurag Ajay, Jiajun Wu, Nima Fazeli, Maria Bauza, Leslie~P Kaelbling, Joshua~B
  Tenenbaum, and Alberto Rodriguez.
\newblock Augmenting physical simulators with stochastic neural networks: Case
  study of planar pushing and bouncing.
\newblock In {\em 2018 IEEE/RSJ International Conference on Intelligent Robots
  and Systems (IROS)}, pages 3066--3073. IEEE, 2018.

\bibitem{masi2021thermodynamics}
Filippo Masi, Ioannis Stefanou, Paolo Vannucci, and Victor Maffi-Berthier.
\newblock Thermodynamics-based artificial neural networks for constitutive
  modeling.
\newblock {\em Journal of the Mechanics and Physics of Solids}, 147:104277,
  2021.

\bibitem{donahue2015long}
Jeffrey Donahue, Lisa Anne~Hendricks, Sergio Guadarrama, Marcus Rohrbach,
  Subhashini Venugopalan, Kate Saenko, and Trevor Darrell.
\newblock Long-term recurrent convolutional networks for visual recognition and
  description.
\newblock In {\em Proceedings of the IEEE conference on computer vision and
  pattern recognition}, pages 2625--2634, 2015.

\bibitem{vinyals2015show}
Oriol Vinyals, Alexander Toshev, Samy Bengio, and Dumitru Erhan.
\newblock Show and tell: A neural image caption generator.
\newblock In {\em Proceedings of the IEEE conference on computer vision and
  pattern recognition}, pages 3156--3164, 2015.

\bibitem{Huang_2018}
Chiou-Jye Huang and Ping-Huan Kuo.
\newblock A deep cnn-lstm model for particulate matter (pm2.5) forecasting in
  smart cities.
\newblock {\em Sensors}, 18(7):2220, Jul 2018.

\bibitem{moriasi2007model}
Daniel~N Moriasi, Jeffrey~G Arnold, Michael~W Van~Liew, Ronald~L Bingner,
  R~Daren Harmel, and Tamie~L Veith.
\newblock Model evaluation guidelines for systematic quantification of accuracy
  in watershed simulations.
\newblock {\em Transactions of the ASABE}, 50(3):885--900, 2007.

\bibitem{willard2020predicting}
Jared Willard, Jordan~Stuart Read, Alison Appling, Samantha Oliver, Xiaowei
  Jia, Paul~C Hanson, Hilary~A Dugan, Robert Ladwig, and Vipin Kumar.
\newblock Predicting water temperature dynamics of unmonitored lake systems
  with meta transfer learning.
\newblock In {\em AGU Fall Meeting Abstracts}, volume 2020, pages H166--0030,
  2020.

\bibitem{chen2021transfer}
Zeng Chen, Huan Xu, Peng Jiang, Shanen Yu, Guang Lin, Igor Bychkov, Alexey
  Hmelnov, Gennady Ruzhnikov, Ning Zhu, and Zhen Liu.
\newblock A transfer learning-based lstm strategy for imputing large-scale
  consecutive missing data and its application in a water quality prediction
  system.
\newblock {\em Journal of Hydrology}, 602:126573, 2021.

\bibitem{kimura2020convolutional}
Nobuaki Kimura, Ikuo Yoshinaga, Kenji Sekijima, Issaku Azechi, and Daichi Baba.
\newblock Convolutional neural network coupled with a transfer-learning
  approach for time-series flood predictions.
\newblock {\em Water}, 12(1):96, 2020.

\bibitem{laptev2018applied}
Nikolay Laptev, Jiafan Yu, and Ram Rajagopal.
\newblock Applied timeseries transfer learning.
\newblock 2018.

\bibitem{liu-et-al:scheme}
Mingliang Liu, Hanqin Tian, Qichun Yang, Jia Yang, Xia Song, Steven~E Lohrenz,
  and Wei-Jun Cai.
\newblock Long-term trends in evapotranspiration and runoff over the drainage
  basins of the gulf of mexico during 1901--2008.
\newblock {\em Water Resources Research}, 49(4):1988--2012, 2013.

\bibitem{lu2013net:kl-one}
Chaoqun Lu and Hanqin Tian.
\newblock Net greenhouse gas balance in response to nitrogen enrichment:
  perspectives from a coupled biogeochemical model.
\newblock {\em Global change biology}, 19(2):571--588, 2013.

\bibitem{Lu2020}
C.~Lu, J.~Zhang, H.~Tian, W.~Crumpton, M.~Helmers, W.~Cai, C.~Hopkinson, and
  S.~Lohrenz.
\newblock {Increased extreme precipitation challenges nitrogen load reduction
  to the Gulf of Mexico}.
\newblock {\em Communications Earth {\&} Environment}, page Accepted, 2020.

\bibitem{yang2015increased}
Qichun Yang, Hanqin Tian, Marjorie~AM Friedrichs, Charles~S Hopkinson, Chaoqun
  Lu, and Raymond~G Najjar.
\newblock Increased nitrogen export from eastern north america to the atlantic
  ocean due to climatic and anthropogenic changes during 1901--2008.
\newblock {\em Journal of Geophysical Research: Biogeosciences},
  120(6):1046--1068, 2015.

\bibitem{yu2019largely}
Zhen Yu, Chaoqun Lu, Hanqin Tian, and Josep~G Canadell.
\newblock Largely underestimated carbon emission from land use and land cover
  change in the conterminous united states.
\newblock {\em Global change biology}, 25(11):3741--3752, 2019.

\bibitem{lu2019-et-al:scheme}
Chaoqun Lu, Jien Zhang, Peiyu Cao, and Jerry~L Hatfield.
\newblock Are we getting better in using nitrogen?: Variations in nitrogen use
  efficiency of two cereal crops across the united states.
\newblock {\em Earth's Future}, 7(8):939--952, 2019.

\bibitem{jones-et-al:scheme}
Christopher~S Jones, Jacob~K Nielsen, Keith~E Schilling, and Larry~J Weber.
\newblock Iowa stream nitrate and the gulf of mexico.
\newblock {\em PloS one}, 13(4):e0195930, 2018.

\bibitem{doi:10.1002/joc.1181}
Timothy~D. Mitchell and Philip~D. Jones.
\newblock An improved method of constructing a database of monthly climate
  observations and associated high-resolution grids.
\newblock {\em International Journal of Climatology}, 25(6):693--712, 2005.

\bibitem{10.1175/BAMS-87-3-343}
Fedor Mesinger, Geoff DiMego, Eugenia Kalnay, Kenneth Mitchell, Perry~C.
  Shafran, Wesley Ebisuzaki, Dušan Jović, Jack Woollen, Eric Rogers,
  Ernesto~H. Berbery, Michael~B. Ek, Yun Fan, Robert Grumbine, Wayne Higgins,
  Hong Li, Ying Lin, Geoff Manikin, David Parrish, and Wei Shi.
\newblock {North American Regional Reanalysis}.
\newblock {\em Bulletin of the American Meteorological Society},
  87(3):343--360, 03 2006.

\end{thebibliography}
\end{document}